\documentclass[10pt,twocolumn,letterpaper]{article}

\usepackage[pagenumbers]{iccv} % To force page numbers, e.g. for an arXiv version

\usepackage{algorithm}% http://ctan.org/pkg
\usepackage{algpseudocode}
\usepackage{amsfonts}
\usepackage{multirow}
\usepackage{amssymb} % for \checkmark
\usepackage{graphicx} % for scaling the table if needed

\usepackage{colortbl}
\definecolor{grey1}{gray}{0.85}

\definecolor{iccvblue}{rgb}{0.21,0.49,0.74}
\usepackage[pagebackref,breaklinks,colorlinks,allcolors=iccvblue]{hyperref}

\def\paperID{4318} % *** Enter the Paper ID here
\def\confName{ICCV}
\def\confYear{2025}

\title{DC-TTA: Divide-and-Conquer Framework \\ for Test-Time Adaptation of Interactive Segmentation}

\author{Jihun Kim* \\
KAIST\\
{\tt\small jihun1998@kaist.ac.kr}
\and
Hoyong Kwon*\\
KAIST\\
{\tt\small kwonhoyong3@kaist.ac.kr}
\and
Hyeokjun Kweon*\\
Chung-Ang University\\
{\tt\small hyeokjunkweon@cau.ac.kr}
\and
Wooseong Jeong\\
KAIST\\
{\tt\small stk14570@kaist.ac.kr}
\and
Kuk-Jin Yoon\\
KAIST\\
{\tt\small kjyoon@kaist.ac.kr}
}

\begin{document}
\maketitle
\begin{abstract}
Interactive segmentation (IS) allows users to iteratively refine object boundaries with minimal cues, such as positive and negative clicks. While the Segment Anything Model (SAM) has garnered attention in the IS community for its promptable segmentation capabilities, it often struggles in specialized domains or when handling complex scenarios (e.g., camouflaged or multi-part objects). 
To overcome these challenges, we propose \textbf{DC-TTA}, a novel test-time adaptation (TTA) framework that adapts SAM on a per-sample basis by leveraging user interactions as supervision. 
Instead of forcing a single model to incorporate all user clicks at once, DC-TTA partitions the clicks into more coherent subsets, each processed independently via TTA with a separated model. 
This \textbf{Divide-and-Conquer} strategy reduces conflicts among diverse cues and enables more localized updates. 
Finally, we merge the adapted models to form a unified predictor that integrates the specialized knowledge from each subset. Experimental results across various benchmarks demonstrate that DC-TTA significantly outperforms SAM’s zero-shot results and conventional TTA methods, effectively handling complex tasks such as camouflaged object segmentation with fewer interactions and improved accuracy.
The code will be available soon.
\end{abstract}    
\section{Introduction}\label{sec:intro}

Interactive segmentation (IS) has emerged as a powerful approach for delineating objects in images with minimal user effort. 
Unlike fully automated methods, IS allows users to iteratively refine segmentation by providing positive or negative clicks, guiding the model toward better results.

\begin{figure}[t]
    \centering
    \includegraphics[width=0.99\linewidth]{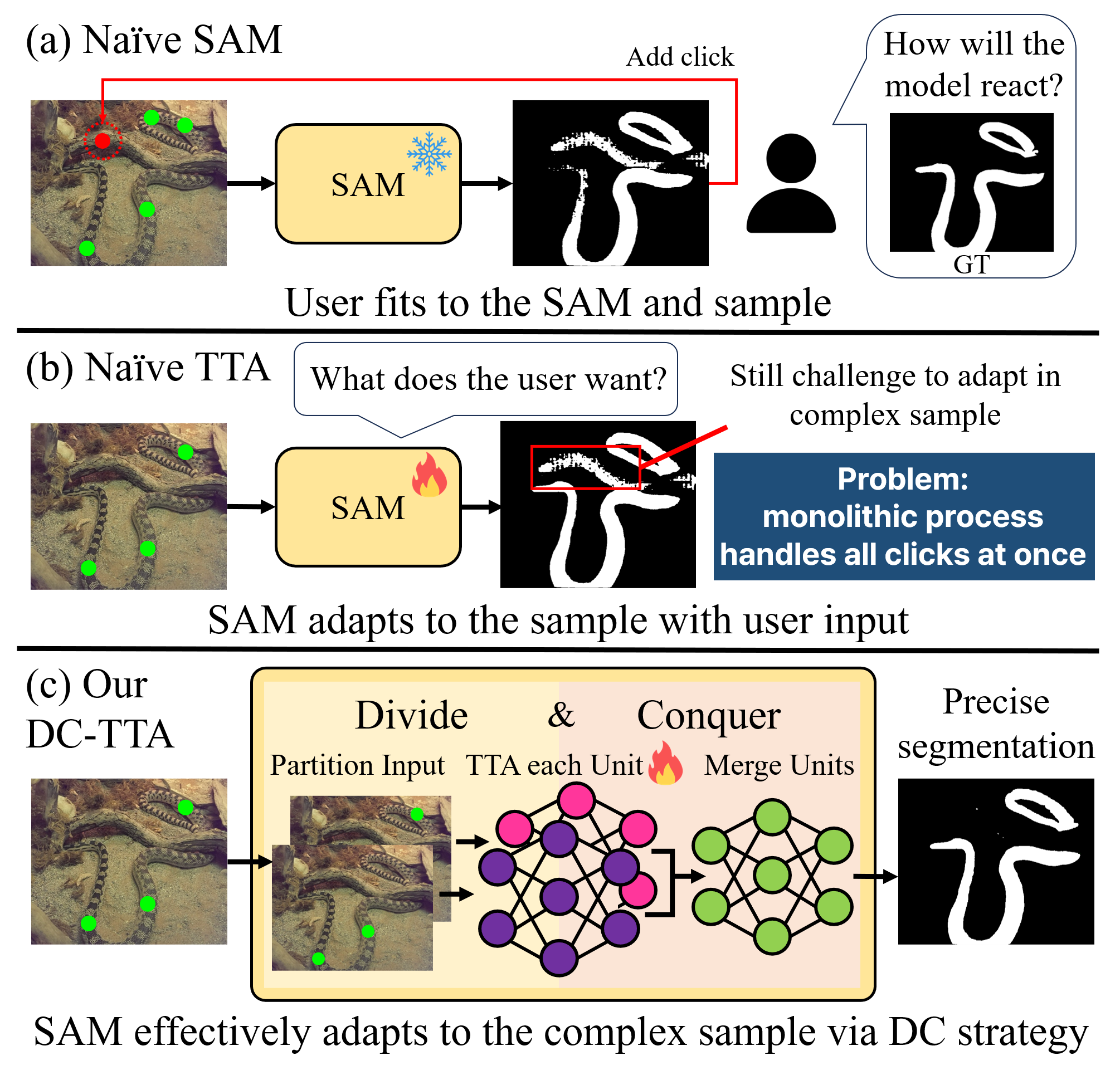}
    \vspace{-6pt}
    \caption{Motivation of Our DC-TTA. (a) Naïve SAM struggles with complex cases. (b) Standard TTA for SAM fails to fully leverage user clicks, as it processes all clicks at once in a monolithic manner. (c) Our DC-TTA adopts a Divide-and-Conquer strategy, effectively handling clicks by updating specialized units for coherent click sets before merging them into a unified segmentation.}
    \label{fig:intro}
    \vspace{-5pt}
\end{figure}

Meanwhile, Segment Anything Model (SAM)~\cite{sam1} recently has shown remarkable zero-shot segmentation capability. 
Given that SAM is inherently designed to take advantage of user-provided prompts, there has been a growing interest in exploiting SAM's capabilities for IS applications~\cite{huang2024focsam, yuan2024open, cheng2023sam, adaptsam}. 
However, while SAM performs well in straightforward cases, its effectiveness diminishes when dealing with complex structures (\textit{e.g.}, camouflaged or multi-part) as in Fig.~\ref{fig:intro}a.
In these challenging scenarios, often the important use cases of IS, the zero-shot performance falls short, necessitating a more adaptive approach.

One such approach is Test-Time Adaptation (TTA), which refines model parameters at inference time leveraging user inputs.
For example, as in Fig.~\ref{fig:intro}b, we update the model by using the new mask, obtained from incorporating the new click, as a pseudo ground truth (GT) for the previous mask without that click.
While we observe that this naive TTA method enhances performance to some extent, it still struggles to effectively integrate the diverse information provided by sequential user clicks.

From our perspective, a key challenge in the naive TTA for SAM within the context of IS is the monolithic process, which attempts to update a single model to integrate and adapt to all user clicks at once. 
While the set of all clicks collectively defines the correct segmentation mask, each click introduces distinct—and sometimes even conflicting—information that can be difficult for a pre-trained SAM to incorporate simultaneously. 
As a result, this all-clicks-at-once approach can capture some global context, but it remains suboptimal when handling the complicated details inherent in IS, where precise updates are required.

To address this challenge, we propose \textbf{DC-TTA}, a novel TTA framework that enhances SAM for IS based on the \emph{\textbf{D}ivide-and-\textbf{C}onquer} (\textbf{DC}) strategy.
As shown in Fig.~\ref{fig:intro}c, rather than updating a single model with all user clicks simultaneously, our method partitions the click set into multiple, localized segmentation units (\underline{\textbf{Divide}}).
Each unit is responsible for a coherent subset of clicks, and the newly received click is assigned to an existing unit or a newly created unit, depending on the overlap with the corresponding mask. 
In addition, we retain a global unit that aggregates all clicks, similar to naive TTA, ensuring that the overall context is preserved and no critical information is overlooked.

Each of these segmentation units has its own model, which is updated independently through TTA, focusing on adapting to the specific clicks associated with that unit (\underline{\textbf{Conquer}}). 
Specifically, when a new click is assigned to a unit, the unit uses it to infer an updated segmentation mask and refine its model. 
While this process mirrors naive TTA, it is performed within each unit  separately, adapting the unit's model only to the assigned subset of clicks.
This approach enables more focused updates to the unit-specific model, ensuring robust adaptation to the relevant user input, while avoiding conflicts from other clicks.

Finally, to integrate the diverse adaptations from different units, we employ a model merging strategy~\cite{ilharco2022editing} via task vectors. 
By combining the localized adaptations of each unit with the broader global context provided by the global unit, the merged model benefits from both detailed, region-specific refinements and the overall context.

We conduct extensive evaluations on eight IS benchmarks, including challenging scenarios such as camouflaged object segmentation. 
The results demonstrate that DC-TTA consistently outperforms all existing methods with substantial margins, including SAM's zero-shot inference and combinations of SAM and TTA approaches.
Additionally, thorough ablation studies show that the DC strategy achieves substantial performance improvements even without the TTA process, highlighting the key advantages of our approach. 
Furthermore, we verify that DC-TTA enhances performance not only when applied to SAM but also to conventional IS methods.
This underscores the broad and adaptable potential of DC-TTA for the field of IS.
\section{Related Works}
\label{sec:rw}

% 지금 최대 고민이 IS에 SAM 쓴걸 2.1이랑 2.2 중 어디에 넣을까...
% 분량 보고 넣는게 나을지도

\subsection{Interactive Segmentation (IS)}
Interactive segmentation (IS) aims to achieve high-quality object segmentation with minimal user interaction. Before the advent of deep learning, early IS methods~\cite{grady2006random,rother2004grabcut,boykov2001interactive, gulshan2010geodesic} relied on optimization-based models. With the advancement of deep learning, various approaches~\cite{liew2017regional, is_li2018interactive, is_ritm, forte2020getting, kontogianni2020continuous, jang2019interactive, lin2020interactive, liu2022pseudoclick, sofiiuk2020f} have been proposed to improve IS performance by integrating user inputs and prior segmentation masks for iterative refinement.
FocalClick~\cite{is_focalclick}, FCFI~\cite{is_fcfi}, and CFR-ICL~\cite{is_cfricl} introduce specialized modules to improve local segmentation quality, while MFP~\cite{is_mfp} extends this idea by incorporating probability maps to improve accuracy. GPCIS~\cite{is_gpcis} leverages Gaussian processes for efficient and robust performance, whereas CPlot~\cite{is_cplot} applies optimal transport theory to better capture diverse user intentions.
Besides, GraCo~\cite{is_graco} and SegNext~\cite{is_segnext} incorporate granularity control and diverse prompts, respectively, to enhance adaptability. 
More recently, Segment Anything Model (SAM)~\cite{sam1} has significantly advanced IS by enabling high generalizability across various domains. 
To further enhance IS performance with the capabilities of SAM, several studies~\cite{huang2024focsam, yuan2024open, cheng2023sam} have been proposed.

\subsection{Test-Time Adaptation (TTA)}
Test-Time Adaptation (TTA)~\cite{colomer2023adapt, wang2023dynamically, volpi2022road, tent, cotta, auxadapt_wacv, auxadapt_cvpr, shot_pmlr20, sar_iclr23, DSS_wacv24, fleuret2021uncertainty, tesla_cvpr23, adacontrast} is the process of adapting a pre-trained model to a target test domain—typically comprising a limited set of test data—in an unsupervised manner, without access to source domain data.
Various online optimization strategies have been explored in TTA research, including entropy minimization~\cite{shot_pmlr20, tent, fleuret2021uncertainty, sar_iclr23}, pseudo-labeling~\cite{shot_pmlr20, fleuret2021uncertainty, tesla_cvpr23, DSS_wacv24}, and contrastive learning~\cite{adacontrast}.
In the IS domain, several studies~\cite{kontogianni2020continuous, lenczner2020interactive} have adopted TTA techniques to further enhance segmentation performance in the target domain by leveraging user clicks as supervision. In recent studies, AdaptSAM~\cite{adaptsam} introduces a framework that incorporates result masks as pseudo-labels, enabling more effective adaptation of the segmentation model during test time.
However, they mainly have focused on adapting the model during monolithic process of IS, unlike our novel DC strategy that allows specialized adaptation on coherent information and merging them into unified segmentation.

% Test-time adaptation (TTA)~\cite{colomer2023adapt,wang2023dynamically,volpi2022road, tent, cotta, auxadapt_wacv, auxadapt_cvpr, shot_pmlr20, sar_iclr23, DSS_wacv24, fleuret2021uncertainty, tesla_cvpr23, adacontrast} adapts a pre-trained model to a target test domain—typically a small set of test data—in an unsupervised manner without access to source domain data. 
% In TTA research, various online optimization strategies have been investigated, including entropy minimization~\cite{shot_pmlr20, tent, fleuret2021uncertainty, sar_iclr23}, pseudo-labeling~\cite{shot_pmlr20, fleuret2021uncertainty, tesla_cvpr23, DSS_wacv24}, and contrastive learning~\cite{adacontrast}.
% However, these methods predominantly target classification tasks, with a few recent studies~\cite{colomer2023adapt,wang2023dynamically,volpi2022road} exploring TTA for image semantic segmentation.

% Continuous adaptation for interactive object segmentation by learning from corrections.
% - IS TTA
% Interactive learning for semantic segmentation in earth observation.
% - IS TTA

\subsection{Model Merging}
Model merging~\cite{yang2024model} combines parameters from multiple fine-tuned models into a unified network without requiring access to the original training data. This approach is particularly useful when fine-tuning large models is computationally prohibitive.
Model merging techniques can be categorized into \textit{Pre-Merging} and \textit{During Merging}. Pre-Merging methods~\cite{ortiz2024task, tang2023parameter, jin2024fine, liu2023tangent, ilharco2022editing} refine models beforehand to improve parameter compatibility, while During Merging methods~\cite{demir2024adaptive, wortsman2022model, wortsman2022robust} focus on the merging process itself. A key approach in this domain is task arithmetic~\cite{ilharco2022editing}, which utilizes task vectors—defined as the difference between the parameters of a fine-tuned model and its pre-trained counterpart—to manipulate model behavior and achieve desired outcomes. Ties-Merging~\cite{yadav2023ties} addresses parameter conflicts through a trim, elect-sign, and merge strategy.
\section{Methods}

This paper proposes DC-TTA, a novel Divide-and-Conquer TTA framework for adapting SAM to IS task.
Before describing our DC-TTA, we first introduce how SAM can be incorporated into the IS process (Sec.~\ref{sec:sam_is}), and describe a naive TTA approach for adapting it (Sec.~\ref{sec:naive_tta}).
These will serve as the foundation upon which we build our divide-and-conquer strategy (Sec.~\ref{sec:dc} and Sec.~\ref{sec:dc_tta}).

\subsection{Interactive Segmentation with SAM}\label{sec:sam_is}

IS aims to iteratively refine an object’s mask through minimal user input. 
In our baseline, we leverage SAM as the segmentation engine. 
Given an input image $I \in \mathbb{R}^{H \times W \times 3}$ and a sequence of user clicks $\{c_t\}$, SAM produces a segmentation mask.
Here, each click $c_t = (x_t, y_t, s_t)$ provided by user at the $t$-th iteration of the IS process consists of spatial coordinates $(x_t, y_t)$ and $s_t \in \{0, 1\}$ indicating a positive ($1$, foreground) or negative ($0$, background) cue.
% Positive clicks guide SAM to expand the mask towards the intended object, while negative clicks refine the mask by excluding unwanted regions.

To condition SAM for segmentation, we define the input prompt at iteration $t$ as $\left( \mathcal{C}_{1:t}, M_{t-1} \right)$, where $\mathcal{C}_{1:t} = \{c_1, c_2, \ldots, c_t\}$ is the set of clicks accumulated up to iteration $t$ and $M_{t-1}\in \{0,1\}^{H \times W}$ is the segmentation mask from the previous iteration.
Hence, the inference of SAM is
\begin{equation}\label{equ:sam}
M_t = \text{SAM}(I;\mathcal{C}_{1:t}, M_{t-1};\theta_0),
\end{equation}
where $M_t$ is the predicted mask at iteration $t$ and $\theta_0$ denotes the pre-trained parameters of SAM.
For $t=1$, we initialize the input prompt for SAM as $\left(\{c_1\}\right)$ without any mask, unless the user provides an initial guess. 

\subsection{Naive Test-Time Adaptation (TTA) Approach}\label{sec:naive_tta}

\begin{figure}[t]
    \centering
    \includegraphics[width=0.99\linewidth]{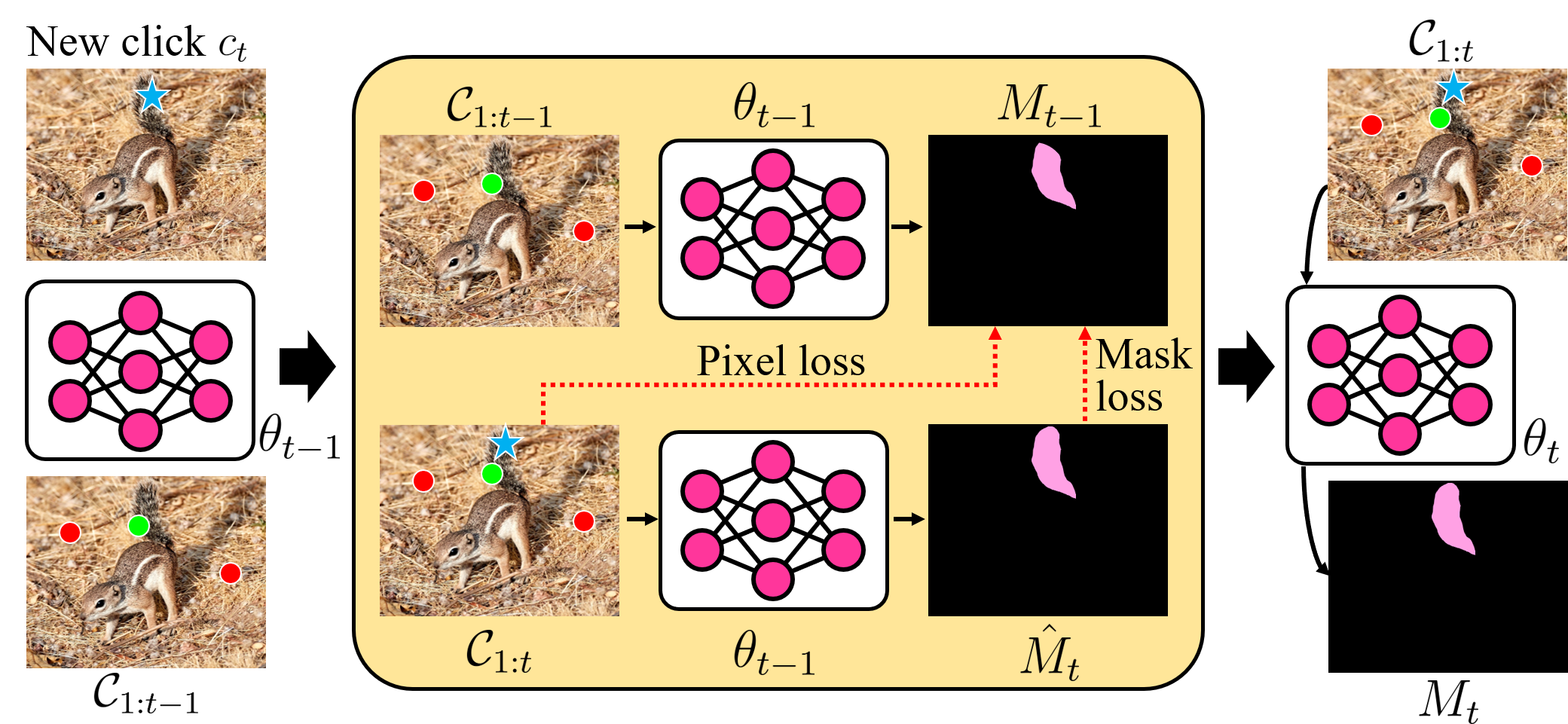}
    \vspace{-10pt}
    \caption{Overview of the naive TTA approach. Each iteration, a new user click $c_t$ is additionally used with previous clicks $\mathcal{C}_{1:t-1}$ to generate an updated mask $\hat{M}_t$. This supervises the previous mask $M_{t-1}$ via a mask-level loss, while a pixel-level loss enforces correctness at the clicked location. The model parameters are then updated once via gradient descent to produce the final mask $M_t$.}
    \vspace{-10pt}
    \label{fig:naive_tta}
\end{figure}

Although SAM is trained on large-scale datasets (SA-1B)~\cite{sam1} and shows strong zero-shot performance, its segmentation quality often degrades in specialized domains, such as cases involving camouflaged objects.
To address this, we propose DC-TTA, a novel TTA framework based on Divide-and-Conquer strategy, which aims to refine SAM per test sample.
Before describing our framework, we first explain a straightforward TTA approach for adjusting SAM to the IS process by incorporating user clicks, as in Fig.~\ref{fig:naive_tta}.

The new click $c_t$ provided at iteration $t$ introduces novel information not present in the previous prediction $M_{t-1}$.
Therefore, the mask of current iteration $\hat{M}_t = \text{SAM}(I;\mathcal{C}_{1:t}, M_{t-1};\theta_{t-1})$, which is inferred incorporating the new click $c_t$, can serve as a refined estimate of the segmentation, with respect to the previous mask $M_{t-1}$.
Here, $\theta_{t-1}$ denotes the parameters of SAM at the previous iteration $t-1$, before the adaptation at iteration $t$. 
Hence, $\hat{M}_t$ is an intermediate output for the $t$-th iterations' TTA process, and we use the $\hat{\;}$ notation on $\hat{M}_t$ to represent this.
We visualize each inference process in the TTA block of Fig.~\ref{fig:naive_tta}.

To adjust the model parameters so that the previous prediction $M_{t-1} = \text{SAM}(I;\mathcal{C}_{1:t-1}, M_{t-2};\theta_{t-1})$ becomes more aligned with the information provided by the new click $c_t$ (and $\hat{M}_t$), we define a mask loss term as the binary cross-entropy loss ($l_{\text{BCE}}$) between $M_{t-1}$ and $\hat{M}_t$.
In addition, we incorporate a pixel loss at the exact clicked pixel, ensuring that the predicted mask aligns with the user’s explicit foreground/background label and thereby reinforcing reliable, high-confidence supervision.
The pixel loss is
\begin{equation}\label{eq:loss_pixel}
\mathcal{L}_{\text{pixel}}(M_{t-1},\mathcal{C}_{1:t}) = \sum_{c_i \in \mathcal{C}_{1:t}} 
l_{\text{BCE}}\Bigl(M_{t-1}(x_i, y_i), s_i \Bigr),
\end{equation}
where $c_i = (x_i, y_i, s_i)$ as we have mentioned.

Accordingly, our loss function for TTA is defined as 
\begin{equation}\label{eq:loss_tta}
\begin{split}
\mathcal{L}_{\text{TTA}}(M_{t-1}, \hat{M}_t, \mathcal{C}_{1:t}) = & l_{\text{BCE}}(M_{t-1}, \hat{M}_t) \\
&+ \mathcal{L}_{\text{pixel}}(M_{t-1},\mathcal{C}_{1:t}).
\end{split}
\end{equation}

The SAM parameters are then updated as
\begin{equation}\label{eq:gd_update}
\theta_t \leftarrow \theta_{t-1} - \eta \nabla_{\theta}\mathcal{L}_{\text{TTA}}(M_{t-1}, \hat{M}_t, \mathcal{C}_{1:t}), 
\end{equation}
where $\eta$ is the learning rate.
With the adapted parameters $\theta_t$, the model re-infers the mask of the current $t$-th iteration:
\begin{equation}\label{eq:tta_inference}
M_t = \text{SAM}(I; \mathcal{C}_{1:t}, M_{t-1}; \theta_t).
\end{equation}

% While this simple TTA scheme provides some improvement, it still suffers when faced with challenging segmentation tasks because forcing a single model to be conditioned by all user clicks—especially when the clicks encode complicated mask information—can be too drastic.

\begin{figure}[t]
    \centering
    \includegraphics[width=0.99\linewidth]{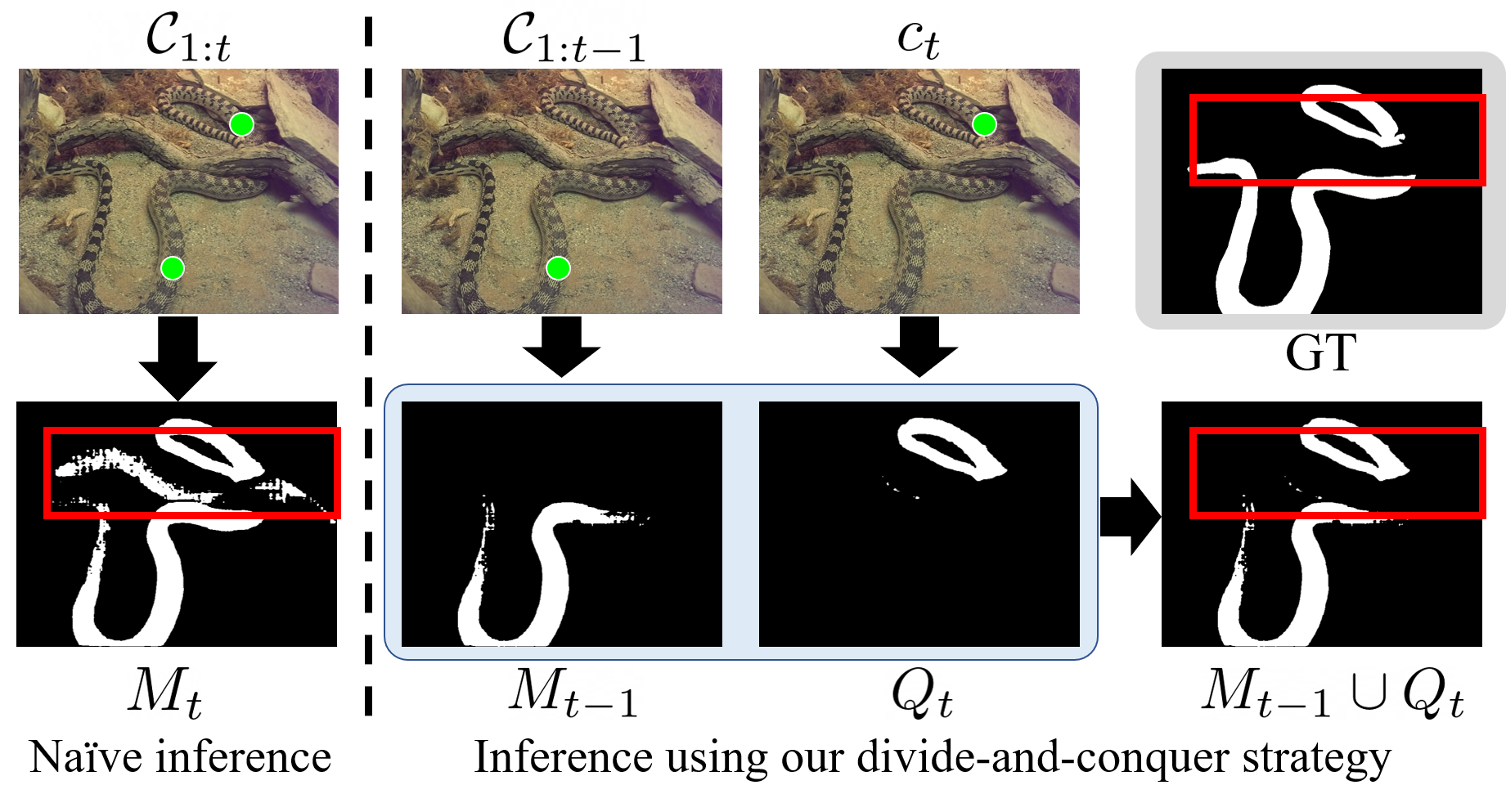}
    \vspace{-5pt}
    \caption{Illustration of our DC strategy (without TTA). Given a new click $c_t$, directly incorporating it into the existing segmentation may not robustly capture new information due to conflicts with prior clicks, resulting in undesirable false positives in background regions. Instead, generating a separate mask using $c_t$ and merging it with the original prediction results in significantly improved segmentation. This shows the effectiveness of our DC strategy for SAM in IS, even without additional learning.}
    \label{fig:dc}
    \vspace{-10pt}
\end{figure}

\subsection{Divide-and-Conquer (DC) Strategy}\label{sec:dc}

In standard IS process, the model is required to process all user clicks at once to perform segmentation. 
This global approach facilitates capturing the object’s overall structure and generally performs well in straightforward cases. 
However, it can be problematic when clicks encode diverse or conflicting information—such as corresponding to regions with complex boundaries or multiple distinct parts—which may lead to ambiguous or suboptimal segmentation.

To address this, we introduce our Divide-and-Conquer (DC) strategy, which partitions the entire set of user clicks into multiple, more coherent subsets. 
The core philosophy of our approach is to group highly correlated clicks together, infer separate masks for each subset, and then aggregate these masks to obtain the final segmentation result.
This allows the model to more effectively resolve conflicts between clicks with different characteristics while still leveraging the unique information provided by each click.

Figure~\ref{fig:dc} illustrates an example of this inference-only IS process (without TTA). Compared to the mask at iteration $t-1$, the newly provided click $c_t$ introduces a positive signal about a previously unidentified region. 
In some cases, directly incorporating $c_t$ into the existing segmentation does not robustly integrate this new information due to conflicts with previous positive clicks. 
In contrast, using only $c_t$ (along with existing negative clicks) yields a separate mask $Q_t$ that better captures the new positive signal. 
Combining this refined mask with the original prediction leads to a improved segmentation result, effectively leveraging the information both from $\mathcal{C}_{1:t-1}$ and the new click $c_t$. 
Quantitatively, we observe that our DC strategy leads to a significant improvement in IS performance, even without any additional learning (refer to DC-only in Table~\ref{tab:main}).

Interestingly, beyond merely improving inference, our DC strategy also reduces confusion during the TTA process.
A key challenge in TTA for IS is updating a single model via a monolithic process, which attempts to incorporate varied information from all clicks simultaneously. This often overwhelms the model, leading to suboptimal performance in complex segmentation tasks.
Our DC strategy stabilizes training by reducing conflicts between signals from different clicks and enhancing SAM’s ability to capture fine-grained details.
Intuitively, our approach resembles employing specialized segmentation experts for different regions of the input image, whose insights are then merged to produce a robust overall segmentation.

\subsection{Divide-and-Conquer TTA (DC-TTA)}\label{sec:dc_tta}

\begin{figure*}[t]
    \centering    \includegraphics[width=0.99\linewidth]{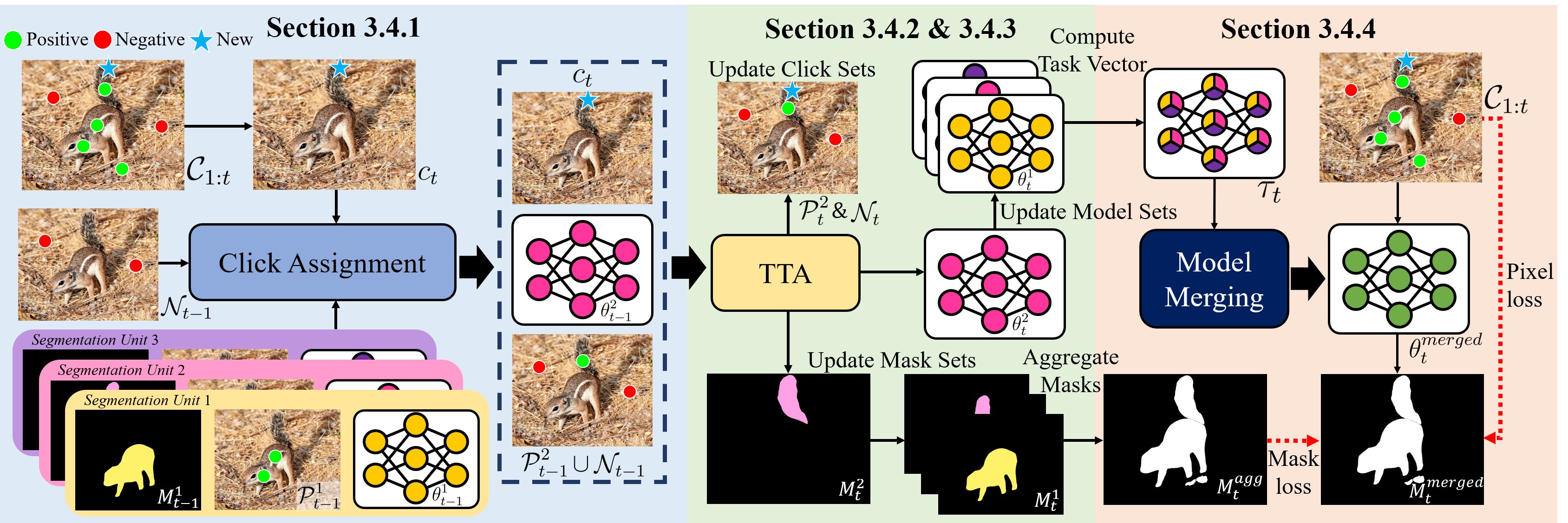}
    \vspace{-8pt}
    \caption{Overview of the proposed DC-TTA framework. First, user clicks are assigned into segmentation units (left). Each unit is then adapted via TTA, resulting in the update of the associated mask and model (center) . Finally, the masks and models of all units are aggregated in pixel and parameter level, respectively (right). This consolidates each unit’s specialized knowledge into a unified segmentation.}
    \label{fig:dc_tta}
\end{figure*}

Figure~\ref{fig:dc_tta} shows the overview of our DC-TTA.
The core philosophy of our DC-TTA approach is to partition the full set of user clicks into more coherent subsets. 
For each subset, we independently perform TTA using a dedicated model.
To formalize this process, we introduce the concept of \emph{Segmentation Units} (units), each encapsulating a set of positive clicks, its predicted mask, and adapted model parameters.
For example, the $k$-th unit is defined as:
\begin{equation}\label{eq:unit}
U^k = \bigl(\mathcal{P}^k,M^k,\theta^k\bigr),
\end{equation}
where $\mathcal{P}^k$ denotes the subset of positive clicks assigned to the $k$-th unit, $M^k$ is the predicted mask for those clicks, and $\theta^k$ represents the model parameters adapted for that unit.

In our DC-TTA framework, each unit is specialized in its own subset of positive clicks, with its model parameters updated accordingly to improve the local mask prediction. 
Aditionally, to capture the overall context of the IS task, we maintain a global segmentation unit ($U^0$) that encompasses all positive clicks regardless of their coherence.
This global unit plays essentially the same role as the baseline IS.

Meanwhile, we define a set $\mathcal{N}$ that contains all negative clicks. 
While positive clicks guide the model to expand the mask towards intended objects, negative clicks serve to exclude undesired regions from the segmentation rather than explicitly indicating specific areas.
Thus, there is no corresponding mask or model for $\mathcal{N}$; instead, it serves as auxiliary constraints uniformly applied to each unit.

\subsubsection{Assigning the New Click to Units}

The DC-TTA process begins when a new click $c_t$ is received. 
If the click is negative (\textit{i.e.}, $s_t=0$), we simply update the set of negative clicks as:
\begin{equation}
\mathcal{N}_t \leftarrow \mathcal{N}_{t-1} \cup \{ c_t \}.
\end{equation}
Then, we apply the TTA process for all units (including $U^0$) using the updated $\mathcal{N}_t$, following the procedure described in Sec.~\ref{sec:naive_tta}. 
This allows each unit to effectively incorporate the information from the newly added negative click.

In the case of a positive click (\textit{i.e.}, $s_t=1$), we evaluate its compatibility with existing segmentation units $U^k$ formed in previous iterations.
Specifically, we assess whether the new click shares characteristics with the masks already estimated by these units or if it introduces novel information that necessitates the creation of a new unit.

To measure this compatibility, we first obtain a mask prediction using the new positive click $c_t$ and the set of negative clicks $\mathcal{N}_{t-1}$ as follows:
\begin{equation}\label{equ:cs_mask}
Q_t = \text{SAM}\Bigl(I; \{c_t\} \cup \mathcal{N}_{t-1}; \theta_0\Bigr).
\end{equation}
The resulting mask $Q_t \in \{0,1\}^{H \times W}$ serves as an estimate of the region indicated by $c_t$. 
Meanwhile, for each unit $k$, its previous mask $M^k_{t-1}$ is computed as:
\begin{equation}\label{equ:unit_mask}
M^k_{t-1} = \text{SAM}\Bigl(I; \mathcal{P}^k_{t-1} \cup \mathcal{N}_{t-1},\, M^k_{t-2}; \theta^k_{t-1}\Bigr).
\end{equation}
We then measure the overlap between $Q_t$ and $M^k_{t-1}$ using the Intersection over Union (IoU):
\begin{equation}\label{equ:iou}
\text{IoU}(Q_t, M^k_{t-1}) = \frac{|Q_t\cap M^k_{t-1}|}{|Q_t\cup M^k_{t-1}|}.
\end{equation}

If the IoU exceeds 0 (\textit{i.e.}, there is overlap), we consider $c_t$ compatible with unit $U^k$ and assign it accordingly.
Note that, regardless of this check, $c_t$ is always added to the global unit ($U^0_t$) that encompasses all clicks.
Since no negative click is added, we keep $\mathcal{N}t \leftarrow \mathcal{N}{t-1}$.

After the assignment, we perform TTA within each affected unit to incorporate the information of the new click, as described in Sec.~\ref{sec:add}.
If there is no overlap, we create a new unit comprising only $c_t$, and adapt it as described in Sec.~\ref{sec:new}.
For a detailed explanation on assigning new points, see Fig.~\textcolor{iccvblue}{A1} in the \textit{Supp}.

\subsubsection{TTA of the Existing Units (including Global Unit)}\label{sec:add}

Once a new positive click $c_t$ is assigned to an existing segmentation unit $U^k = (\mathcal{P}^k, M^k, \theta^k)$, we perform TTA on that unit’s model. 
This process is similar to Sec.~\ref{sec:naive_tta}, except it is performed within the unit independently.
We first update the unit's positive click set as:
\begin{equation}
\mathcal{P}^k_t \leftarrow \mathcal{P}^k_{t-1} \cup \{c_t\}.
\end{equation}
Then, as in Eq.~\ref{equ:unit_mask}, we compute the previous mask prediction $M^k_{t-1}$ for unit $k$. 
This mask is obtained using the positive click set $\mathcal{P}^k_{t-1}$, the negative click set $\mathcal{N}{t}$, and the model parameters $\theta^k{t-1}$.
To guide the update of $M^k_{t-1}$ during TTA, we compute an intermediate mask incorporating the newly assigned click $c_t$ as:
\begin{equation}
\hat{M}^k_t = \text{SAM}\Bigl(I; \, \mathcal{P}^k_{t} \cup \mathcal{N}_{t},\, M^k_{t-1};\, \theta^k_{t-1}\Bigr).
\end{equation}
Using $\hat{M}^k_t$ as a pseudo GT for the unit, we update the unit-specific model parameters via gradient descent:
\begin{equation}\label{equ:unit_update}
\theta^k_t \leftarrow \theta^k_{t-1} - \eta \nabla_\theta \, \mathcal{L}_{\text{TTA}}(M^k_{t-1}, \hat{M}^k_t, \mathcal{P}_{t}^k\cup\mathcal{N}_{t})
\end{equation}

With the updated parameters $\theta^k_t$, the adapted model produces an updated mask for unit $k$:
\begin{equation}\label{equ:unit_inference}
M^k_t = \text{SAM}\Bigl(I; \, \mathcal{P}^k_{t} \cup \mathcal{N}_{t},\, M^k_{t-1};\, \theta^k_t\Bigr).
\end{equation}

\subsubsection{TTA of the Newly Created Unit}\label{sec:new}

If the new positive click $c_t$ cannot be assigned to any existing unit (except global unit), we create a new segmentation unit dedicated to $c_t$.
We denote the index of this new unit as $K+1$, where $K$ is the current number of existing units (excluding the global unit).
We initialize its model parameters from the original SAM parameters ($\theta^{K+1}_{\text{init}} \leftarrow \theta_0$) and set its positive click set to $\mathcal{P}_t^{K+1} \leftarrow { c_t }$.

Additionally, we adapt the new unit's model parameter via TTA using the set of negative clicks $\mathcal{N}_{t}$ as auxiliary cues. 
Specifically, we first obtain 
\begin{equation}
M^{K+1}_{\text{init}} = \text{SAM}\Bigl(I; \, \mathcal{P}_t^{K+1};\, \theta^{K+1}_{\text{init}}\Bigr),
\end{equation}
using only the positive click, without negative clicks.
We then update $\theta^{K+1}_{\text{init}}$, by using $Q_t$ (refer to Eq.~\ref{equ:cs_mask}) and $\mathcal{N}_{t}$ to supervise $M^{K+1}_{\text{init}}$ as 
\begin{equation}
\theta^{K+1}_t \leftarrow \theta^{K+1}_{\text{init}} - \eta \nabla_\theta \, \mathcal{L}_{\text{TTA}}(M^{K+1}_{\text{init}}, Q_t, \mathcal{P}_t^{K+1}\cup\mathcal{N}_t).
\end{equation}

Using the updated parameters $\theta^{K+1}t$, the mask $M^{K+1}{t}$ for the new unit is computed following the same procedure as other units.
This ensures that, although the new unit is initiated solely by the positive click, its mask prediction also reflects the constraints imposed by the negative clicks. 
The newly created unit $U^{K+1} = (\mathcal{P}^{K+1}_t, M^{K+1}_t, \theta^{K+1}_t)$ is then maintained for subsequent iterations.

\subsubsection{Integration via model merging}\label{sec:mm}

After performing unit-wise TTA, we integrate the outputs from the units to obtain the unified segmentation result. 
In our DC-TTA framework, each unit $U^k$ produces its own adapted mask $M^k_t$, computed using the unit-specific parameters $\theta^k_t$ and the corresponding subset of positive clicks $\mathcal{P}^k_t$, along with negative clicks $\mathcal{N}_t$. 
We aggregate these masks via a pixel-wise union:
\begin{equation}\label{equ:mask_aggregate}
M^{\text{agg}}_t = \bigcup_{k=0}^{K} M^k_t,
\end{equation}
where index $0$ corresponds to the global unit encompassing all clicks, and $K$ is the total number of non-global units. 
This union ensures that $M^{\text{agg}}_t$ integrates the various localized segmentation information from each unit.

Integrating masks via this simple pixel-wise union directly combines localized segmentation outputs, indeed improving performance.
However, it still does not fully leverage the rich adaptation information captured within each unit’s model. 
For example, each unit-specific model is expected to encode fine-grained, localized features through tailored TTA. 
These nuanced representations are reflected in differences between adapted and original SAM parameters. 
Motivated by task vectors in the field of multi-task learning, we find that merging the unit-specific models at the parameter level can consolidate this specialized knowledge from all units into a unified model.
Specifically, for each unit, the corresponding task vector can be defined as:
\begin{equation}\label{equ:task_vector}
\tau^k_t = \theta^k_t - \theta_0,
\end{equation}
which captures the adaptation induced by the clicks in $k$-th unit, relative to the original SAM parameters $\theta_0$. 
We perform model merging by combining these task vectors with the original parameters $\theta_0$ via element-wise addition :
\begin{equation}\label{equ:merged_model}
\theta^{\text{merged}}_t = \theta_0 + \gamma\tau^0_t + \gamma^2\sum_{k=1}^{K} \tau^k_t,
\end{equation}
where $\gamma$ is a scaling parameter for each task vector.

To stabilize the merged model, we perform an additional fine-tuning step.
To this end, we employ the aggregated mask $M^{\text{agg}}_t$ in Eq.~\ref{equ:mask_aggregate} as pseudo GT. 
The merged model's prediction is obtained as:
\begin{equation}
M^{\text{merged}}_t = \text{SAM}(I; \mathcal{C}_{1:t}, M_{t-1}; \theta^{\text{merged}}_t).
\end{equation}

Then, $\theta^{\text{merged}}_t$ is updated via gradient descent:

\begin{equation}\label{equ:final_update}
\theta_t \leftarrow \theta^{\text{merged}}_t - \eta \nabla_\theta  
\mathcal{L}_{\text{TTA}}(M^{\text{merged}}_t, M^{\text{agg}}_t, \mathcal{C}_{1:t})
\end{equation}
where $\theta_t$ is the final model parameter of iteration $t$.

Finally, the refined segmentation mask $M_t$ is as:
\begin{equation}\label{equ:merged_inference}
M_t = \text{SAM}(I; \mathcal{C}_{1:t}, M_{t-1}; \theta_t).
\end{equation}

This integrated approach—combining the union of unit predictions with task vector-based model merging and a final fine-tuning step—enables our DC-TTA framework to efficiently learn from diverse, localized clicks, thereby overcoming the limitations of a monolithic adaptation process.

\section{Experimental Results}\label{sec:exp}

\begin{table*}[t]
    \caption{Performance comparison on eight IS benchmarks (NoC and FR at 85\% and 90\% IoU with 20 clicks). The proposed DC-TTA consistently achieves lower NoC and FR values than the baselines, demonstrating its effectiveness in handling challenging IS tasks.}
    \vspace{-8pt}
    \label{tab:main}
    \centering
    \small
    \setlength{\tabcolsep}{2pt}
    \begin{tabular}{l c c c c c c c c c c c c c c c c}
        \toprule
        \multirow{2}{*}{Method} & \multicolumn{4}{c}{CAMO~\cite{camo}} & \multicolumn{4}{c}{COD10k~\cite{cod10k}} & \multicolumn{4}{c}{TRASHCAN~\cite{trashcan}} & \multicolumn{4}{c}{ISTD~\cite{istd}} \\
        \cmidrule(lr){2-5} \cmidrule(lr){6-9} \cmidrule(lr){10-13} \cmidrule(lr){14-17}
        %  & \multicolumn{2}{c}{20@85} & \multicolumn{2}{c}{20@90} & \multicolumn{2}{c}{20@85} & \multicolumn{2}{c}{20@90} & \multicolumn{2}{c}{20@85} & \multicolumn{2}{c}{20@90} & \multicolumn{2}{c}{20@85} & \multicolumn{2}{c}{20@90} \\
        % \cmidrule(lr){2-3} \cmidrule(lr){4-5} \cmidrule(lr){6-7} \cmidrule(lr){8-9} \cmidrule(lr){10-11} \cmidrule(lr){12-13} \cmidrule(lr){14-15} \cmidrule(lr){16-17}
        % & NoC & FR & NoC & FR & NoC & FR & NoC & FR& NoC & FR& NoC & FR & NoC & FR & NoC & FR\\
        & NoC85 & FR85 & NoC90 & FR90 & NoC85 & FR85 & NoC90 & FR90 & NoC85 & FR85 & NoC90 & FR90 & NoC85 & FR85 & NoC90 & FR90\\
        \midrule
        % Example rows (Replace these with actual values)
        Baseline~\cite{sam1} & 6.88 & 12.40 & 10.45 & 31.20 & 7.88 & 24.93 & 11.43  & 45.21 & 9.06 & 26.06 & 13.30 & 52.29 & 9.33& 27.17 & 11.57 & 41.02\\
        TENT~\cite{tent} & 6.80 & 12.40 & 10.42 & 32.80 & 7.89 & 25.27 & 11.44 & 45.06 & 9.08 & 26.21 & 13.31 & 52.43 & 9.29 & 27.06 & 11.57 &  40.53\\
        AdaptSAM~\cite{adaptsam} & 6.77 & 11.60 & 10.25 & 30.00 & 7.78 & 24.14 & 11.28 & 43.88 & 8.83 & 23.51 & 13.08 & 49.33 & 9.18& 25.94 & 11.41 & 39.30\\
        DC-only & 6.72 & 11.60 & 10.20 & 30.4 & 7.55 & 22.90 & 11.11 & 42.50 & 9.01 & 25.35 & 13.24 & 51.26 & 8.90& 22.99 & 11.12 & 35.78\\
        Ours & \textbf{6.41} &\textbf{ 9.60} & \textbf{9.58} & \textbf{25.60} & \textbf{7.42} & \textbf{21.62} &\textbf{10.80} & \textbf{39.73} & \textbf{7.89} & \textbf{17.13} & \textbf{12.05} & \textbf{38.80} & \textbf{8.66} & \textbf{21.44} & \textbf{10.88}& \textbf{34.87}\\ 
        \midrule
        \multirow{2}{*}{Method} & \multicolumn{4}{c}{Berkeley~\cite{berkeley}} & \multicolumn{4}{c}{DAVIS~\cite{davis}} & \multicolumn{4}{c}{COCOMVal~\cite{cocomval}} & \multicolumn{4}{c}{PascalVOC~\cite{pascalvoc}} \\
        \cmidrule(lr){2-5} \cmidrule(lr){6-9} \cmidrule(lr){10-13} \cmidrule(lr){14-17}
        % & \multicolumn{2}{c}{20@85} & \multicolumn{2}{c}{20@90} & \multicolumn{2}{c}{20@85} & \multicolumn{2}{c}{20@90} & \multicolumn{2}{c}{20@85} & \multicolumn{2}{c}{20@90} & \multicolumn{2}{c}{20@85} & \multicolumn{2}{c}{20@90} \\
        % \cmidrule(lr){2-3} \cmidrule(lr){4-5} \cmidrule(lr){6-7} \cmidrule(lr){8-9} \cmidrule(lr){10-11} \cmidrule(lr){12-13} \cmidrule(lr){14-15} \cmidrule(lr){16-17}
        % & NoC & FR & NoC & FR & NoC & FR & NoC & FR& NoC & FR& NoC & FR & NoC & FR & NoC & FR\\
        & NoC85 & FR85 & NoC90 & FR90 & NoC85 & FR85 & NoC90 & FR90 & NoC85 & FR85 & NoC90 & FR90 & NoC85 & FR85 & NoC90 & FR90\\
        \midrule
        % Example rows (Replace these with actual values)
        Baseline~\cite{sam1} & 1.80 & 0.00 & 2.41 & 1.00 & 4.21 & 9.57 & 5.38 & 15.07 & 3.01 & 3.25 & 4.84 & 10.38 & 2.64 & 1.81 & 3.12 & 3.04\\
        TENT~\cite{tent} & \textbf{1.77} & 0.00 & 2.41 & 1.00 & 4.22 & 9.57 & 5.39  & 15.36 & 3.01 & 3.13 & 4.90 & 11.00 & 2.64& 1.81 & 3.11 & 2.99 \\
        AdaptSAM~\cite{adaptsam} & \textbf{1.77} & 0.00 & 2.29 & \textbf{0.00} & 4.16 & 9.28 & 5.39 & 15.07 & 2.97 & 2.88 & 4.71 & 9.25 & 2.62 & 1.70 & 3.08 & 2.69\\
        DC-only & 1.79 & 0.00 & 2.32 & 1.00 & 4.11 & 9.28 & 5.36 & 15.07 & 3.02 & 3.38 & 4.78 & 10.13 & 2.59 & 1.55 & 3.06 & 2.63\\
        Ours & 1.80 & 0.00 & \textbf{2.17} & \textbf{0.00} & \textbf{4.07} & \textbf{8.99} & \textbf{5.26}& \textbf{15.07}&\textbf{2.71} &\textbf{0.88} &\textbf{4.22}  & \textbf{5.00} &\textbf{2.54} & \textbf{1.08}& \textbf{2.99} & \textbf{1.99}\\
        % Add more rows as necessary
        \bottomrule
    \end{tabular}
    \vspace{-10pt}
\end{table*}

\begin{figure*}[t]
    \centering
    \includegraphics[width=0.99\linewidth]{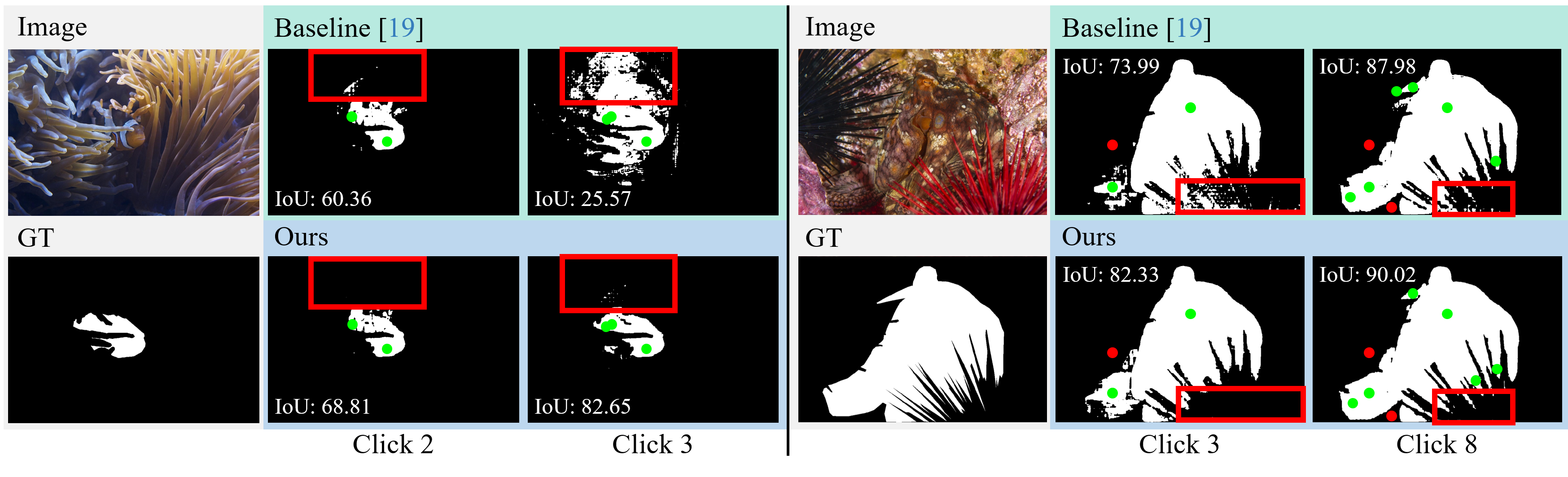}
    \vspace{-10pt}
    \caption{A qualitative comparison between the baseline method and our DC-TTA.
Each subfigure shows the segmentation result after a certain number of user clicks (green for positive, red for negative), along with the resulting IoU.
While the baseline SAM suffers from false positives (in red boxes) as additional positive user clicks are added, our DC-TTA effectively mitigates this issue by utilizing the divide-and-conquer strategy, resulting in robust segmentation with fewer errors. }
    \label{fig:qual}
    \vspace{-10pt}
\end{figure*}

\begin{figure*}[t]
    \centering
    \includegraphics[width=0.85\linewidth]{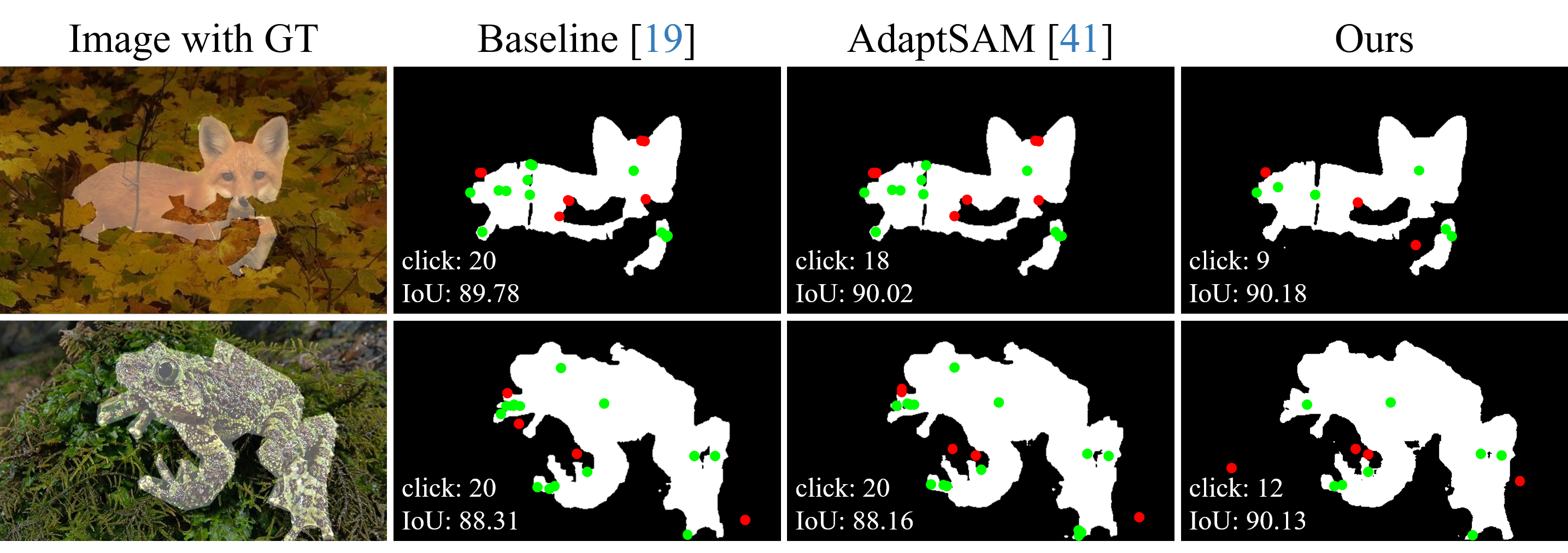}
    \vspace{-10pt}
    \caption{Qualitative examples of challenging camouflaged objects from CAMO~\cite{camo} dataset. Despite the high camouflage and complex backgrounds, our DC-TTA framework successfully identifies the objects, achieving high IoU with relatively few clicks.}
    \vspace{-10pt}
    \label{fig:fail_succ}
\end{figure*}

\subsection{Settings}
\noindent\textbf{Datasets:}
% 우리 실험한 각 데이터셋들 아주 간략하게만 정리
% 우리는 2개의 camouflage dataset CAMO, COD10k를 포함하여 8개의 dataset에서 our method를 evaluation하였다. CAMO와 COD10k는 각각 250, 2026 images로 이루어져 있으며 foreground objet와 background의 구분이 어려운 camouflage image로 이루어져 있다. TRASHCAN은 train set과 test set을 병합하여 evaluate하였으며, 이미지 내의 TRASH 카테고리를 포함한 5062개의 image를 대상으로 TRASH 객체에 대한 segmentation을 평가하였다. ISTD는 train set과 test set을 병합하여 1870개의 그림자 image로 이루어졌다. 
% Berkeley는 100개의 instance가 포함된 96개의 이미지로 구성되었다. DAVIS는 50개의 비디오로부터 추출된 345개의 frame과 high-quality mask로 이루어졌다. 
% COCOMval: f-BRS: Rethinking Backpropagating Refinement for Interactive Segmentation가 제작.
% COCOMVal은 MS COCO 2017 validation set에서 class별 10개씩 총 800개의 image로 구성되었다. PascalVOC는 1449개 image의 3427개의 instances로 구성되었다. CAMO, COD10k, TRASHCAN은 한개의 image에 여러개의 instance가 있는경우 함께 목표하도록 하였으며, 나머지 데이터셋은 각 instance마다 evaluation이 진행되었다. 
We evaluate DC-TTA on eight IS benchmarks. 
CAMO~\cite{camo} and COD10K~\cite{cod10k} consist of 250 and 2,026 images camouflaged images, respectively.
For TRASHCAN~\cite{trashcan}, we use 5,062 underwater object images containing trash objects from its train/test sets.
ISTD~\cite{istd} consists of 1,870 images for shadow segmentation, and both train and test set are used. 
Berkeley~\cite{berkeley} consists of 96 images containing 100 instances in total. 
DAVIS~\cite{davis} includes 345 frames extracted from 50 videos. %, along with high-quality masks. 
COCOMVal~\cite{cocomval} is a subset of the MS COCO 2017~\cite{coco} validation set, containing 800 images with 10 images per class.
PascalVOC~\cite{pascalvoc} consists of 1,449 images with 3,427 instances. 
For CAMO, COD10K, and TRASHCAN, multiple instances in a single image are evaluated together. 
For the other datasets, evaluation is conducted on a per-instance basis.

\noindent\textbf{Metrics:}
% NoC, FR
% 우리는 previous works [1,2,3]과 같은 click simulation strategy를 따랐으며, prediction과 ground truth를 비교하여 가장 큰 에러 영역의 중심에 next click이 수행된다. 
% 우리는 정해진 Intersection over Union (IoU)에 도달하기까지 필요한 평균 number of clicks를 표현하는 Number of Clicks (NoC)를 평가 지표로 이용하였으며, target IoU를 \mathcal{T}라고 할때 NoC\mathcal{T}로 표시하였다.
% 따로 표기하지 않는 한 maximum click 수는 20회이며, maximum click까지 target IoU를 도달하지 못한 sample의 비율을 Failure Rate (FR)를 이용하였다. 
We follow the click simulation strategy used in previous works~\cite{is_li2018interactive, is_ritm, is_focalclick, is_simpleclick}, where the next click is placed at the center of the largest error region based on the comparison between the prediction and the GT.
For evaluation, we use Number of Clicks (NoC), which measures the average number of clicks required to reach a predefined IoU threshold $\mathbf{T}$, denoted as NoC$\mathbf{T}$.
Failure Rate (FR) represents the percentage of samples that fail to reach the target IoU within the maximum number of clicks.
Unless otherwise specified, we set the maximum number of clicks to 20.

\noindent\textbf{Implementation details:}
% input image resolution은 SAM의 기본 값인 1024x1024로 설정되었으며, VIT-b backbone의 SAM이 이용되었다. 
% Test time adpation 과정에서 SAM의 prompt encoder와 mask decoder만 AdamW optimizer with learning rate 1e-5로 1 iteration optimize되었다. 
% Click Assignment과정에서 new positive click과 기존의 negative clicks가 함께 입력되어 표현 영역이 판단되었다. 
% 최종 prediction에서 weight parameter $\gamma = 0.7$로 model merging이 수행되었다. 
% our method의 divide and conquer strategy의 유용성을 보이기 위하여 TTA 없이 DC strategy만 수행한 후, 각 cluster의 도출 mask를 aggregate한 실험을 진행하였으며, 이를 DC-only로 나타내었다. 
%
We use SAM~\cite{sam1} with a ViT-B~\cite{vit} backbone as our baseline and an input resolution of $1024 \times 1024$.
For the results of using larger backbones (\textit{e.g.}, ViT-L), please refer to \textit{Supp}.
% In all cases, the second of the three valid masks is selected for the first click, and the masks for subsequent clicks are predicted without using SAM's multi-mask option.
During TTA, only the prompt encoder and mask decoder of SAM are optimized for one iteration using AdamW~\cite{adamw} with $\eta=1e-5$.
Model merging is performed with a scaling parameter of $\gamma=0.7$.

\subsection{Interactive Segmentation Results}

Table~\ref{tab:main} summarizes the comparisons on eight IS benchmarks.
Our method consistently outperforms the existing SAM-based IS approaches, including baseline SAM~\cite{sam1}, TENT~\cite{tent} and AdaptSAM~\cite{adaptsam} across all datasets. 
These results verify that DC-TTA achieves robust and powerful adaptation of SAM for IS task.
Further, we also test DC-only setting without any TTA, where $M^{\text{agg}}_t$ is directly used for the final prediction (see Eq.~\ref{equ:mask_aggregate}).
DC-only still outperforms the baseline, clearly showing the effectiveness of our DC strategy.
In \textit{Supp}., we also provide experimental results under the setting with 95\% IoU threshold and 30 clicks.

We provide a qualitative comparison between the baseline SAM~\cite{sam1} and our DC-TTA. Figure~\ref{fig:qual} shows that DC-TTA incorporates new click information effectively, suppressing erroneous predictions (\textit{e.g.}, false positives in the red boxes). 
This improvement is achieved by performing additional adaptation with the given click information while leveraging our DC strategy to mitigate potential conflicts. 
Consequently, our proposed method attains higher IoU with the same number of clicks.
Further, Fig.~\ref{fig:fail_succ} shows how DC-TTA succeeds in segmenting camouflaged objects, where the baseline SAM~\cite{sam1} and AdaptSAM~\cite{adaptsam} struggled (under FR90 setting). 
Despite the high degree of camouflage, DC-TTA achieves substantial segmentation with fewer clicks. 
\textit{Supp}. provides more visual samples.
% Notably, it avoids the failures that often occur in such challenging scenes, demonstrating the robustness of our divide-and-conquer strategy.

\begin{table}[t]
\caption{Ablation studies of DC-TTA on CAMO~\cite{camo} and TRASHCAN~\cite{trashcan}. MM denotes model merging. We compare the baseline SAM~\cite{sam1} with incremental additions (A: Naive TTA, B: DC-only, C: DC-TTA without MM, and D: our DC-TTA). }
\vspace{-5pt}
\label{tab:abl}
\centering
\setlength{\tabcolsep}{3pt}
\resizebox{0.99\linewidth}{!}{
\begin{tabular}{c|c|cc|cc|cc}
\hline
 & \multirow{2}{*}{TTA} & \multicolumn{2}{c|}{DC} & \multicolumn{2}{c|}{CAMO~\cite{camo}} &  \multicolumn{2}{c}{TRASHCAN~\cite{trashcan}} \\
 &  & Divide & MM & NoC90 & FR90 & NoC90 & FR90\\ 
\hline \hline
Baseline & & & & 10.45 & 31.20 & 13.30 & 52.29\\ \hline
A & \checkmark & & & 10.03 & 28.40 & 12.54 & 43.18 \\ \hline
B & & \checkmark & & 10.20& 30.40 & 13.24 & 51.26\\ \hline
C & \checkmark & \checkmark & & 9.83& 27.20 & 12.29 & 39.67\\ \hline
D & \checkmark & \checkmark & \checkmark & \textbf{9.58} & \textbf{25.60} & \textbf{12.05} & \textbf{38.80}\\ \hline
\end{tabular}
}
\vspace{-8pt}
\end{table}

\begin{figure}[t]
    \centering
    \includegraphics[width=0.99\linewidth]{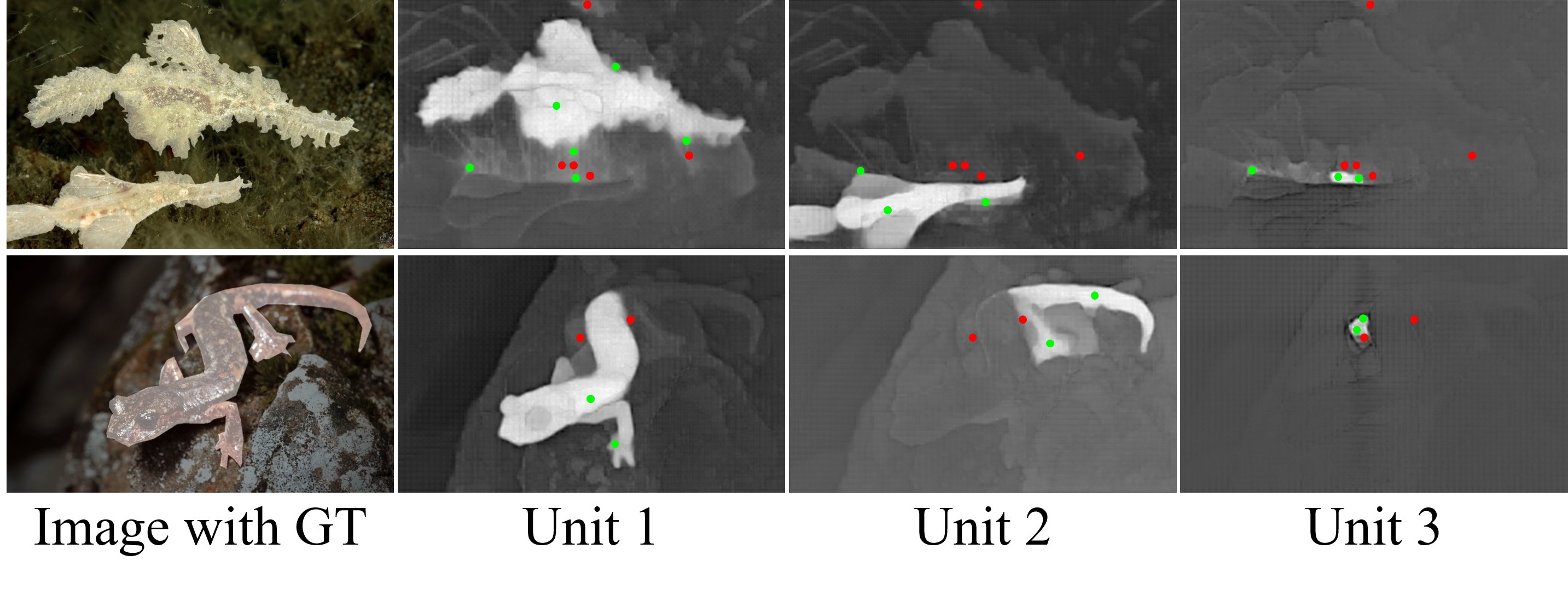}
    \vspace{-10pt}
    \caption{Segmentation units identified in our DC-TTA. User clicks are partitioned into coherent subsets, each yielding a distinct partial prediction aligns with the underlying object structure.}
    \label{fig:unit_partition}
    \vspace{-14pt}
\end{figure}

\subsection{Additional Experiments}
\subsubsection{Ablation Studies}
% For further analysis, we conduct ablation studies of DC-TTA on CAMO~\cite{camo} and TRASHCAN~\cite{trashcan}.
Table~\ref{tab:abl} shows how each component of our DC-TTA framework affects performance.
% The baseline is vanilla SAM (\textit{i.e.}, without TTA or DC). 
Exp A applies naive-TTA in Sec.~\ref{sec:naive_tta} only, while Exp B (DC-only) uses our DC strategy alone. Exp C and Exp D are DC-TTA without and with model merging, respectively.
We observe that each component provides an non-overlapping improvement, where their combination achieves high segmentation performance. 
We also conduct ablation study on scaling parameter $\gamma$ in \textit{Supp}.

\begin{table}[t]
    \caption{Results of applying DC-TTA on the existing IS methods (20-click setting). When integrated with existing IS frameworks (RITM~\cite{is_ritm}, FocalClick~\cite{is_focalclick}, and MFP~\cite{is_mfp}), DC-TTA consistently lowers both NoC90 and FR90.}
    \label{tab:others}
    \vspace{-5pt}
    \centering
    % \small
    % \renewcommand{\arraystretch}{1.2}
    \setlength{\tabcolsep}{3pt}
    \resizebox{0.87\linewidth}{!}{
    \begin{tabular}{l c c c c}
        \toprule
        \multirow{2}{*}{Method} & \multicolumn{2}{c}{CAMO~\cite{camo}} & \multicolumn{2}{c}{COD10k~\cite{cod10k}}  \\
        \cmidrule(lr){2-3} \cmidrule(lr){4-5} 
        & NoC90 & FR90 & NoC90 & FR90 \\
        \midrule
        % Example rows (Replace these with actual values)
        % CDNet~\cite{is_cdnet} & 11.00 & 29.20 & 12.69 & 49.26  \\
        % CDNet+Ours & \textbf{10.64} & \textbf{25.20} & \textbf{12.62} & \textbf{46.69}  \\ 
        RITM~\cite{is_ritm}~$_{\text{ICIP22}}$ & 10.51 & 20.00 & 12.02 & 40.38  \\
        % TENT &  &  &  &  &  &  &  &  \\
        RITM+Ours & \textbf{9.83} & \textbf{18.40} & \textbf{11.52} & \textbf{37.96}  \\
        FocalClick~\cite{is_focalclick}~$_{\text{CVPR22}}$ & 7.66 & 13.20 & 12.26 & 46.84  \\
        FocalClick+Ours & \textbf{7.40} & 13.20 & \textbf{12.20} & \textbf{46.45} \\ 
        MFP~\cite{is_mfp}~$_{\text{CVPR24}}$ & 7.06 & 11.60 & 11.86 & 44.92  \\
        MFP+Ours & \textbf{6.92} & 11.60 & \textbf{11.81} & \textbf{44.27}  \\ 
        \midrule
        \multirow{2}{*}{Method}  & \multicolumn{2}{c}{TRASHCAN~\cite{trashcan}} & \multicolumn{2}{c}{ISTD~\cite{istd}} \\
        \cmidrule(lr){2-3} \cmidrule(lr){4-5} 
        & NoC90 & FR90 & NoC90 & FR90  \\
        \midrule
        % Example rows (Replace these with actual values)
        % CDNet~\cite{is_cdnet} & 12.76 & 44.63 & 11.02 & 33.69  \\
        % CDNet+Ours & \textbf{12.74} & \textbf{39.99} & \textbf{10.24} & \textbf{25.40}  \\ 
        RITM~\cite{is_ritm}~$_{\text{ICIP22}}$ & 11.10 & 31.47 & 10.50 & 29.20  \\
        % TENT &  &  &  &  &  &  &  &  \\
        RITM+Ours & \textbf{10.60} & \textbf{27.82} & \textbf{9.73} & \textbf{24.28}  \\
        FocalClick~\cite{is_focalclick}~$_{\text{CVPR22}}$ & 11.03 & 33.94 & 5.59 & 10.00  \\
        FocalClick+Ours & \textbf{10.16} & \textbf{26.37} & \textbf{5.34} & \textbf{9.41}  \\ 
        MFP~\cite{is_mfp}~$_{\text{CVPR24}}$ & 8.80 & 21.02 & 5.29 & 7.11  \\
        MFP+Ours & \textbf{8.73} & \textbf{20.45} & \textbf{5.13} & \textbf{6.74}  \\ 
        \bottomrule
    \end{tabular}
    }
    \vspace{-15pt}
\end{table}

\subsubsection{DC-TTA with conventional IS methods}
Table~\ref{tab:others} reports the performance of our DC-TTA when integrated with several conventional IS methods on four benchmarks. 
% When combined with CDNet~\cite{is_cdnet}, RITM~\cite{is_ritm}, and FocalClick~\cite{is_focalclick}, our approach consistently reduces both NoC90 and FR90.
When combined with RITM~\cite{is_ritm}, FocalClick~\cite{is_focalclick}, and MFP~\cite{is_mfp}, our approach consistently improves performance.
It demonstrates that DC-TTA not only can be applied for adapting SAM but also improves other IS methods by effectively incorporating user clicks.

\subsubsection{Analysis on DC Strategy}
Figure~\ref{fig:unit_partition} shows the example of how DC-TTA partitions user clicks into distinct segmentation units. 
Each unit, corresponding to a coherent subset of positive clicks, produces its own partial logit and mask prediction. 
The figure demonstrates that the clicks are distributed across units in a manner that aligns with the underlying object structure (multi instances or parts), suggesting that DC-TTA effectively decomposes complex IS tasks into more manageable units.
\vspace{-5pt}

\section{Conclusion}\label{sec:con}
\vspace{-5pt}
We present DC-TTA, a novel divide-and-conquer test-time adaptation framework that partitions user clicks into coherent subsets and adapts the model locally before merging the specialized outputs. By reducing conflicts among diverse clicks and performing more localized updates, our approach outperforms both zero-shot SAM and conventional TTA methods in challenging interactive segmentation scenarios. These results highlight the potential of DC-TTA to handle complex object structures with fewer interactions for more efficient and robust user-driven segmentation.

{
    \small
    \bibliographystyle{ieeenat_fullname}
    \bibliography{main}
}

% WARNING: do not forget to delete the supplementary pages from your submission 
% \input{sec/X_suppl}

\end{document}